\documentclass[lettersize,journal]{IEEEtran}
\usepackage{amsmath,amsfonts}
\usepackage{algorithmic}
\usepackage{algorithm}
\usepackage{array}
\usepackage[caption=false,font=normalsize,labelfont=sf,textfont=sf]{subfig}
\usepackage{textcomp}
\usepackage{stfloats}
\usepackage{url}
\usepackage{verbatim}
\usepackage{graphicx}
\usepackage{cite}
\usepackage[utf8]{inputenc} % allow utf-8 input
\usepackage{hyperref}       % hyperlinks
\usepackage{booktabs}       % professional-quality tables
\usepackage{nicefrac}       % compact symbols for 1/2, etc.
\usepackage{microtype}      % microtypography
\usepackage{xcolor}         % colors
\usepackage{todonotes}
\usepackage{multirow}
\usepackage{colortbl}
\definecolor{mygray}{gray}{0.9}
\newcolumntype{P}[1]{>{\centering\arraybackslash}p{#1}}
\usepackage{wrapfig}
\usepackage{enumitem}
\usepackage[capitalize]{cleveref}

\definecolor{commentcolor}{rgb}{0.2, 0.6, 0.2}
\definecolor{classcolor}{rgb}{0.7, 0.1, 0.2}
\definecolor{functioncolor}{rgb}{0.1, 0.1, 0.8}
\definecolor{keywordcolor}{rgb}{0.6, 0.2, 0.6}
\newcommand{\mycomment}[1]{\textcolor{commentcolor}{#1}}

\newcommand{\mykeyword}[1]{\textcolor{keywordcolor}{\textbf{#1}}}

\begin{document}

\title{Exploiting Discriminative Codebook Prior for Autoregressive Image Generation}

\author{
Longxiang Tang,
Ruihang Chu,
Xiang Wang,
Yujin Han,
Pingyu Wu,
Chunming He,\\
Yingya Zhang,
Shiwei Zhang,
Jiaya Jia,~\IEEEmembership{Fellow,~IEEE}%
\thanks{Corresponding author: Ruihang Chu.}%
\thanks{Longxiang Tang and Jiaya Jia are with the Department of Computer Science and Engineering, The Hong Kong University of Science and Technology, Clear Water Bay, Hong Kong (e-mail: lloong.x@gmail.com; jia@cse.ust.hk).}%
\thanks{Ruihang Chu, Xiang Wang, Yujin Han, Pingyu Wu, Yingya Zhang and Shiwei Zhang are with the Tongyi Lab, Alibaba Group, Hangzhou 311121, China (e-mail: ruihangchu@gmail.com).}%
\thanks{Chunming He is with the Department of Biomedical Engineering, Duke University, Durham, NC 27708 USA.}
\thanks{This work has been submitted to the IEEE for possible publication. Copyright may be transferred without notice, after which this version may no longer be accessible.}
}

\markboth{Submitted to IEEE TPAMI}%
{Tang \MakeLowercase{\textit{et al.}}: Exploiting Discriminative Codebook Prior for Autoregressive Image Generation}


\maketitle

\begin{abstract}
Advanced discrete token-based autoregressive image generation systems first tokenize images into sequences of token indices with a codebook, and then model these sequences in an autoregressive paradigm. While autoregressive generative models are trained only on index values, the prior encoded in the codebook, which contains rich token similarity information, is not exploited.
Recent studies have attempted to incorporate this prior by performing naive k-means clustering on the tokens, helping to facilitate the training of generative models with a reduced codebook.
However, we reveal that k-means clustering performs poorly in the codebook feature space due to inherent issues, including token space disparity and centroid distance inaccuracy.
In this work, we propose the Discriminative Codebook Prior Extractor (DCPE) as an alternative to k-means clustering for more effectively mining and utilizing the token similarity information embedded in the codebook.
DCPE replaces the commonly used centroid-based distance, which is found to be unsuitable and inaccurate for the token feature space, with a more reasonable instance-based distance. Using an agglomerative merging technique, it further addresses the token space disparity issue by avoiding splitting high-density regions and aggregating low-density ones.
Extensive experiments demonstrate that DCPE is plug-and-play and integrates seamlessly with existing codebook prior-based paradigms. With the discriminative prior extracted, DCPE accelerates the training of autoregressive models by 42\% on LlamaGen-B and improves final FID and IS performance.
% Code: \url{http://gitlab.alibaba-inc.com/longxiang/DCPE}
\end{abstract}


\section{Introduction}
\label{sec:intro}

\begin{figure*}[t]
    \centering
    \includegraphics[width=\linewidth]{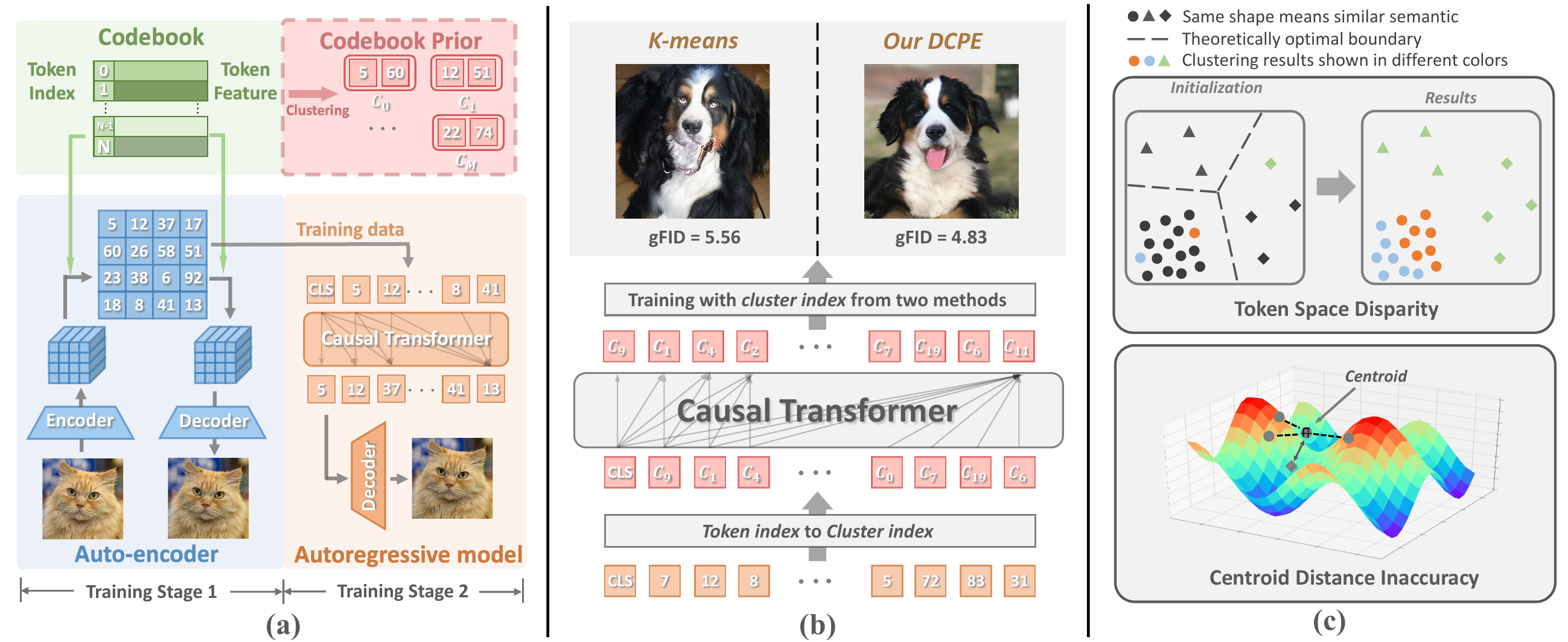} 
    % \vspace{-2mm}
    \caption{\textbf{(a)}: A common framework for discrete token-based autoregressive image generation methods. While the image tokenizer and the autoregressive model are trained separately, we extract token similarity information as a codebook prior to assist the training of the latter. \textbf{(b)}: A toy experiment on training an autoregressive model with cluster indices produced by different clustering algorithms. K-means leads to performance degradation, whereas our DCPE achieves better generation quality. \textbf{(c)}: Illustrations of the token space disparity and centroid distance inaccuracy issues, highlighting the limitations of applying naive k-means clustering to the codebook token feature space.}
    \label{fig1}
    % \vspace{-3mm}
\end{figure*}

\IEEEPARstart{A}{utoregressive} image generation~\cite{esser2021taming,sun2024autoregressive,tian2024visual,zhou2024transfusion,fan2024fluid,li2024autoregressive} has developed rapidly as a new paradigm for image generation, demonstrating increasingly promising performance and the ability to scale up like large language models (LLMs)~\cite{brown2020language,touvron2023llama,bai2023qwen,liu2024deepseek,qu2025does,zhou2024uniqa,zhou2025gamma,liu2025hybrid} compared to traditional diffusion methods~\cite{dhariwal2021diffusion,ho2020denoising,rombach2022high,peebles2023scalable}.
Among them, approaches~\cite{esser2021taming,sun2024autoregressive,tian2024visual,han2024infinity} using discrete tokenizers have garnered increasing attention due to their LLM-aligned structures~\cite{team2024chameleon,wu2024janus}, which enable the realization of a unified world model. 
As shown in \cref{fig1} (a), a standard discrete token-based autoregressive image generation framework consists of two main components:
an image tokenizer~\cite{esser2021taming} and an autoregressive Transformer~\cite{radford2019language,touvron2023llama}. The image tokenizer is composed of an image encoder, a learnable codebook, and an image decoder.
Training images are first fed into the encoder to extract visual features, followed by vector quantization to convert the continuous features into a sequence of index values. This quantization is achieved by identifying the token vector in the codebook that is closest to each visual feature. These index sequences are then used to train the autoregressive Transformer model in a causal modeling manner.
To generate a new image during inference, the well-trained autoregressive model generates appropriate token index sequences, which are then decoded into images by the decoder.

This autoregressive image generation framework, adopted by most existing works~\cite{esser2021taming,sun2024autoregressive,tian2024visual,han2024infinity}, separates the training of the image tokenizer and the autoregressive model, expecting the latter to fit the distribution of image tokens directly from the index sequences. 
However, since the image tokenizer is used in encoding and decoding, it naturally contains rich information that may benefit the training of generation models.
For instance, tokens with similar semantics should ideally share similar logits during inference.
Some recent works~\cite{hu2025improving,guo2025improving} have started to investigate the token similarity prior encoded in the codebook by incorporating a k-means clustering operation on tokens to facilitate the autoregressive models' training. 
For example, IAR~\cite{hu2025improving} encourages the model to correctly predict the cluster where the target token is located through an auxiliary cluster-oriented loss, while CTF~\cite{guo2025improving} introduces a coarse-to-fine pipeline with token clustering, first predicting the cluster indices and then refining them to the fine-grained tokens.

Although some success was achieved, we suggest that the naive k-means clustering algorithm used to extract the codebook prior may not perform well in the token feature space. To demonstrate this, we conduct a toy experiment, as shown in \cref{fig1} (b).
We cluster the 16k tokens from the LlamaGen~\cite{sun2024autoregressive} codebook into 8k clusters. The clustering results are then used to convert the token indices in the training sequences into cluster indices. After training autoregressive models with these cluster indices, we randomly select token within the predicted cluster to enable appropriate image decoding.
The underlying intuition is that an effective clustering algorithm should be able to group tokens with similar semantics into the same cluster. Thus, models are expected to learn to predict accurate cluster and maintain reasonable generation quality. However, results show that models struggle to fit the cluster index sequences produced by naive k-means clustering.

These outcomes reveal that naive k-means clustering shows suboptimal ability on capturing the inherent token similarity information from the codebook. As illustrated in \cref{fig1} (c), we attribute this phenomenon to two primary factors: (1) \textit{Token Space Disparity}. The auto-encoder must handle information at multiple granularities to compress and reconstruct an image; thus, the token feature space becomes non-uniform in density~\cite{van2017neural,williams2020hierarchical,huh2023straightening}. Since k-means initializes clusters randomly and updates them without accounting for density variations, it can inadvertently split high-density regions and group together tokens from sparse regions that are actually dissimilar. (2) \textit{Centroid Distance Inaccuracy}. While the token feature space can be viewed as a manifold within the higher-dimensional space~\cite{beyer1999nearest,angiulli2018behavior,souiden2022survey}, the way k-means calculates distance using cluster centroid, which is the arithmetic mean of vectors, is not ideal. This approach may cause dissimilar tokens to appear close due to a flawed distance measure. These two factors contribute to intra-cluster dissimilarity, and working with such inaccurate clustering results can lead to suboptimal generation performance. We provide more discussion on this in \cref{sec:prelim} and experimental demonstration in \cref{sec:abl}.

In this paper, we introduce the Discriminative Codebook Prior Extractor (DCPE), designed to more effectively exploit the token similarity information embedded in the codebook. Recognizing the token space disparity issue, we employ an agglomerative clustering strategy that iteratively merges the two most similar clusters, proving highly effective in the non-uniform token feature space. This density-aware strategy successfully prevents the splitting of high-density areas and the aggregation of low-density ones. To address the issue of centroid distance inaccuracy, DCPE computes inter-cluster distances using instance-based distance measures, eliminating the dependency on inaccurate high-dimensional centroids. Using DCPE, we are able to train autoregressive models with a clustered codebook, achieving accelerated training and yielding comparable or even better performance. For example, in LlamaGen-B training, acceleration reached 42\% for FID and 55\% for IS. As a plug-and-play solution, DCPE can also be seamlessly integrated into existing codebook prior-based methods, delivering enhanced generation performance.

Our main contributions can be summarized as three-fold:
\begin{itemize}[leftmargin=*]
    \item We investigate and reveal two critical issues in extracting codebook priors for autoregressive image generation: the token space disparity and the centroid distance inaccuracy.
    \item We introduce the DCPE for codebook prior extraction as a replacement for the widely adopted yet flawed k-means method, addressing the identified issues through an agglomerative strategy and instance-based distance calculation.
    \item Experimental results demonstrate that our method is plug-and-play and integrates seamlessly with existing works, offering improved performance with a reduced codebook size.
\end{itemize}

\section{Related Work}

\textit{Autoregressive Image Generation.}
With the rapid development of generative visual architectures such as diffusion models~\cite{dhariwal2021diffusion,ho2020denoising,rombach2022high,peebles2023scalable,he2024diffusion,zhu2024instantswap,wei2025dreamrelation} and autoregressive paradigm~\cite{esser2021taming,sun2024autoregressive,tian2024visual,zhou2024transfusion,fan2024fluid,li2024autoregressive,pang2024next,pang2025randar,jiao2025flexvar,li2025autoregressive,ren2025beyond,xu2025direction,ma2025token,cheng2025tensorar,chen2025tts,hezipar,wu2025alitok}, visual generation has achieved impressive progress and enabled the synthesis of high-fidelity image content.
Among them, the autoregressive paradigm, which predicts the next visual content in a sequence based on previous ones, has recently received considerable attention inspired by the great success in the field of natural language processing~\cite{brown2020language,touvron2023llama,bai2023qwen,liu2024deepseek}.
LlamaGen~\cite{sun2024autoregressive} adapts the ``next-token prediction'' manner based on the Llama~\cite{touvron2023llama} architecture, attempting to autoregressively predict the next visual token. 
MAR~\cite{li2024autoregressive} leverages a diffusion head to model per-token probability, eliminating the need for discrete tokenizers. % and achieving high-quality generation. 
VAR~\cite{tian2024visual} introduces the next scale prediction paradigm, generating visual content in a coarse-to-fine resolution prediction form. 
TiTok~\cite{yu2024image} explores a 1D Transformer-based tokenizer to transform 2D images into a 1D latent sequence by inserting extra learnable tokens.
RAR~\cite{yu2024randomized} permutes raster order token prediction into different factorization orders, enhancing the ability of capturing bidirectional dependencies.
In this work, we focus on discrete token-based autoregressive image generation and aim to incorporate the prior encoded in the codebook to enhance generation capabilities.

\textit{Codebook Manipulation.}
While image generation methods using discrete tokenizers have attracted increasing attention, several studies have begun exploring ways to manipulate the codebook to improve or accelerate the training of generative models.
HyperHill~\cite{li2023resizing} enhances the codebook vectors with co-occurrence-aware hyperbolic embeddings and reorders them via a Hilbert curve to enable flexible resizing of the codebook, improving efficiency and reconstruction quality. 
CVQ-VAE~\cite{zheng2023online} mitigates codebook collapse by using encoded features as anchors to update inactive code vectors, thereby improving codebook utilization and integration with existing architectures.
RAQ~\cite{seo2024raq} enables VQ-based generative models to support multiple bitrates by adapting codebooks in a data-driven manner without requiring retraining.
While these approaches focus on adjusting the codebook, they do not incorporate any codebook information into the training process of generative models. In contrast, our work aims to extract prior knowledge from the codebook for the training of generative models, offering a plug-and-play integration with existing methods.

\textit{Training with Codebook Prior.}
Existing autoregressive image generation frameworks with discrete tokenizers typically separate the training of the image tokenizer and the generative model. Recent studies~\cite{hu2025improving,guo2025improving,qiu2025robust} have begun to explore how to leverage token similarity information derived from the codebook to improve generation quality.
IAR~\cite{hu2025improving} proposes a codebook rearrangement strategy with balanced k-means clustering and a cluster-oriented cross-entropy loss that guides the model to correctly predict the cluster where the target token is located, enhancing the generation quality and robustness.
CTF~\cite{guo2025improving} introduces a coarse-to-fine pipeline, which trains an autoregressive model that sequentially predicts coarse labels obtained by clustering, and an auxiliary model that predicts fine-grained labels conditioned on their coarse labels.
RobustTok~\cite{qiu2025robust} proposes training the tokenizer with perturbed tokens, which enhances the robustness of tokenizer thus alleviates the discrepancy of reconstruction and generation qualities.
These approaches typically extract codebook priors using naive k-means clustering or top-k selection without analyzing the underlying attributes of token feature distribution.
In this work, we address this gap by proposing an extractor that is more suitable for the token feature space, thereby exploiting a more discriminative codebook prior.

\section{Method}

\subsection{Preliminary}
\label{sec:prelim}

\subsubsection{Autoregressive Image Generation} We first formulate the training and inference pipeline of most existing discrete token-based autoregressive image generation methods~\cite{esser2021taming,sun2024autoregressive,hu2025improving}. Given an image tokenizer consisting of an encoder $E$, a decoder $D$ and a codebook $\mathcal{Z}$, we can transform an image $x\in\mathbb{R}^{H\times W\times3}$ into an index sequence which is used to train the autoregressive models. First, the image feature $z$ is extracted by the encoder $E$: $z=E(x)\in\mathbb{R}^{h\times w\times d}$, where $d$ represents the dimension of each feature vector. Then we apply vector quantization to the image features with a codebook $\mathcal{Z}=\{v_1, v_2, \cdots, v_N\}$, where $v_i\in\mathbb{R}^{d}$ is the $i$-th token vector and $N$ is the codebook size. Specifically, for each feature vector $z^{(i,j)}\in\mathbb{R}^{d}$ in the image feature, we look for the token index $q^{(i,j)}$ of its nearest token neighbor in the codebook:
\begin{equation}
    q^{(i,j)} = \arg \min \limits_{v_k\in \mathcal{Z}} \| z^{(i,j)} - v_k \| \in [0,N)
\end{equation}
where $v_k$ is the $k$-th vector in the codebook $\mathcal{Z}$, and $\| \cdot \|$ denotes the commonly used Euclidean distance. After obtaining the quantized index sequence $z^q=\{q^{(i,j)}\}\in\mathbb{Z}^{h\times w}$ from all training images, we can use them to train a autoregressive generation model $\mathcal{G}$ with a next-token prediction paradigm: $\hat{z}^q_i=\mathcal{G}(\hat{z}^q_1, \hat{z}^q_2, \cdots, \hat{z}^q_{i-1})$. During inference, the well-trained autoregressive model generates a token sequence $\hat{z}^q\in\mathbb{Z}^{h\times w}$ in a causal manner, which is then decoded into an image by the decoder as $\hat{x}=D(\hat{z}^q, \mathcal{Z})$.

\subsubsection{Codebook Clustering with K-means} Here we formulate the widely used k-means clustering algorithm on codebook tokens. The goal of codebook clustering is to assign cluster indices $\{c_i\}_{i=1}^{k}$ to each token $v_i$ in the codebook, where $k < N$ is the number of clusters. The k-means algorithm maintains a centroid set $\{\mu_i\}_{i=1}^{k}$ and assigns each token to the cluster with the nearest centroid. At the beginning, centroids $\{\mu_i^{(0)}\}_{i=1}^{k}$ are initialized using $k$ randomly selected token vectors in the codebook, and the initial cluster assignments $\{c_i^{(0)}\}_{i=1}^{k}$ can be obtained by: $c_i^{(0)} = \arg \min_{j \in [0,k)} \| \mu_j^{(0)} - v_i \|$, where the superscripts in parentheses represent different iterations. After assigning the cluster indices to each token in the current iteration, k-means then updates the centroids by computing the arithmetic mean of all tokens in the same cluster, denoted by $\mu_i^{(1)} = \frac{1}{|C_i^{(0)}|} \sum_{v_j \in C_i^{(0)}} v_j$, where $C_i^{(0)} = \{ v_j | c_j^{(0)} = i \}$. This process of updating the cluster indices and centroids continues iteratively until convergence.

\subsubsection{Discussion of K-means Clustering on Codebook} Despite its widespread use, the k-means algorithm is not well-suited for clustering in the token feature space here. This limitation primarily stems from the sparse distribution of token vectors in the high-dimensional space. K-means algorithm may group token vectors with variant semantics into the same cluster, resulting in poor intra-cluster consistency~\cite{assent2012clustering,ikotun2023k,na2010research}. We attribute this issue to two main factors in our context: token space disparity and centroid distance inaccuracy. First, because the codebook encodes information at multiple granularities, the token feature space exhibits non-uniform density~\cite{van2017neural,williams2020hierarchical,huh2023straightening,tang2023consistency,fang2024real}. K-means algorithm does not take this variation into account and updates all clusters simultaneously, which may lead to dense token regions being split, while semantically distinct tokens in sparse regions are grouped together. Furthermore, as token vectors lie on a complex manifold in high-dimensional space, their arithmetic mean, i.e. the cluster centroid, often falls outside the manifold~\cite{beyer1999nearest,angiulli2018behavior,souiden2022survey}. This misalignment results in inaccurate distance calculations between centroids and token vectors. These challenges highlight the need for clustering methods better suited to the structure of codebook tokens.

\subsection{Discriminative Codebook Prior Extractor}
\label{sec:DCPE}

The aforementioned issues of token space disparity and centroid distance inaccuracy in k-means clustering can result in the token clusters containing mixed semantic information, which provides inaccurate codebook prior for the subsequent training of the autoregressive model~\cite{hu2025improving,guo2025improving}. To address these challenges, we propose the Discriminative Codebook Prior Extractor (DCPE) to better exploit the token similarity information embedded within the codebook.

The main reason k-means fails to handle the token feature space disparity issue is that it updates all clusters at the same time without priority. This approach does not guarantee that regions with higher token density are merged prior to those with lower density. Moreover, the issue is further exacerbated by the error-prone random initialization strategy. To minimize the risk of high-density tokens being separated and low-density regions being merged, we adopt the concept from agglomerative clustering~\cite{lukasova1979hierarchical,ackermann2014analysis} to design a deterministic clustering algorithm. Our method prioritizes merging the two most similar clusters, ensuring that the clustering result is sensitive to token density. Specifically, we initialize all token vectors $\{v_i\}_{i=1}^{N}$ as individual clusters, i.e., $C^{(0)}_i = \{v_i\}$, and then perform $N-k$ merge operations to obtain $k$ clusters. Each time, we greedily merge the two most similar token clusters, as expressed by the following formula:
\begin{equation}
    \label{eq:our_merge}
    C^{(i+1)}_s = C^{(i)}_s \cup C^{(i)}_t,\ \text{where}\ s,t=\arg\min \mathcal{D}(C^{(i)}_s,C^{(i)}_t)
\end{equation}
where $i=1, 2,\cdots, N-k$ represents the iteration, and $\mathcal{D}(\cdot)$ is the distance measure between two clusters. This strategy, by design, ensures that clusters from high-density areas are merged before those from low-density areas, effectively preventing the occurrence of clusters containing tokens with divergent semantics, which would hinder the training of autoregressive models.  Meanwhile, this design requires more computation compared to heuristic algorithm k-means, which will be discussed in the following section.

Considering the discussion regarding the issue of centroid distance inaccuracy, it is inappropriate to use the centroid distance to measure the inter-cluster similarity in the token feature space. Therefore, we decide to eliminate the use of centroids and instead rely solely on valid inter-token distances to indirectly measure the distance between clusters. In this regard, DCPE adopts a simple yet effective instance-based inter-cluster distance approach. For two clusters $C_s=\{v_i\}_{i=0}^{n_s}$ and $C_t=\{v_j\}_{j=0}^{n_t}$, where $n_s$ and $n_t$ represent the number of tokens in each cluster, we first calculate all the pairwise distances between the tokens in the two clusters: $\hat{\mathcal{D}}(v_i, v_j)=\| v_i-v_j \|$. The inter-cluster distance is then obtained by averaging all these instance-level distances, which is expressed as:
\begin{equation}
    \label{eq:our_dis}
    \mathcal{D}(C_s,C_t) = \mathbb{E}_{v_i\sim C_s, v_j\sim C_t}[\hat{\mathcal{D}}(v_i, v_j)]
\end{equation}

This distance is used in \cref{eq:our_merge} to replace the widely used centroid distance in existing methods. This approach shares a similar underlying idea with unweighted average linkage agglomerative clustering~\cite{nei2000molecular}, and performs effectively in the token feature space, improving the discrimination of clusters and thereby enhancing the subsequent training of generative models with better codebook prior.

\subsection{Implementation and Analysis}
\label{sec:impl_anal}

According to \cref{eq:our_dis}, the calculation of inter-cluster distance in our DCPE requires computing every pairwise distance between tokens in the two clusters. If the algorithm is implemented naively based on this formula, the computational complexity becomes quite high, reaching $O(N^3d)$ (proof is provided in the appendix). Moreover, due to the frequent tensor slicing involved, the approach fails to fully leverage GPU acceleration. This issue becomes particularly severe when the codebook size is large~\cite{esser2021taming,sun2024autoregressive}.

\begin{figure}[t]
    \centering
    \includegraphics[width=\linewidth]{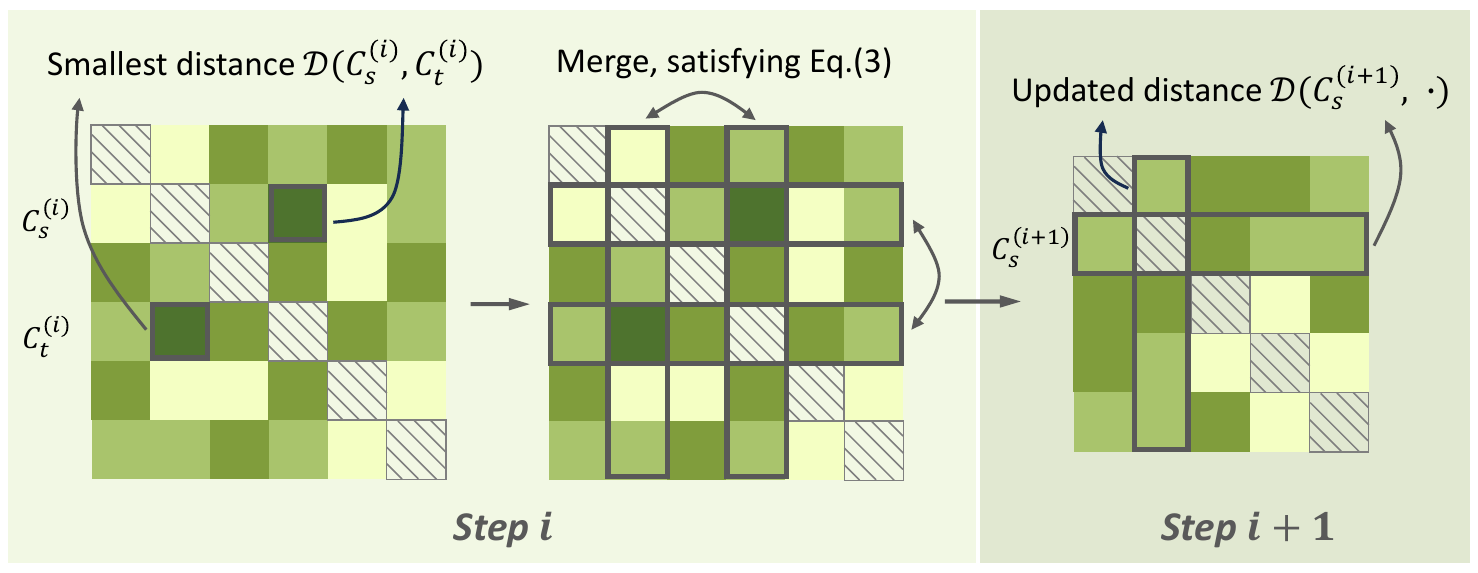} 
    % \vspace{-2mm}
    \caption{Illustration of the implementation of our DCPE. Each grid represents the distance between two clusters (darker color indicates smaller distance). We maintain inter-cluster distances and update the distance matrix during agglomerative clustering, enabling accelerated computation.}
    \label{fig2}
    % \vspace{-3mm}
\end{figure}

To improve the efficiency of our DCPE and thereby enhance its practical applicability, we explore implementation-level optimizations. Given that DCPE employs an agglomerative clustering strategy, it is straightforward to observe that the distances between clusters not involved in the merging process do not need to be recomputed; they can be directly inherited from the previous iteration. However, for newly formed clusters resulting from a merge, computing their distances to all remaining clusters still involves substantial tensor slicing operations, leading to computational inefficiency. Fortunately, the instance-based cluster distance measure we adopt can be interpreted as a combination of inter-token distances. This allows us to propagate new distances with existing ones: averaging the distances between each of the two original clusters and the other clusters. With this optimization, we only need to maintain an inter-cluster distance matrix, which is initialized with pairwise token distances and incrementally updated throughout the clustering process, thereby alleviating the burden of redundant computation. An illustration of the implementation is shown in \cref{fig2}, and the detailed algorithmic steps are provided in \cref{algo:pseduo_code}.
 
By maintaining and updating the inter-cluster distance matrix, we significantly improved the efficiency of our DCPE. As shown in the pseudo-code, the complexity of the optimized algorithm is reduced to $O(N^3)$. This not only lowers the theoretical complexity but also leads to significant improvements in the constant factors of the time complexity. Furthermore, the optimized implementation is well-suited to parallel computation and can fully leverage GPU acceleration. Experimental results demonstrate that, for a codebook size of $16k$ in Llamagen~\cite{esser2021taming}, our optimized implementation reduces the processing time from 4.4 hours in the naive implementation to only 39.6 seconds. Compared to the hundreds of GPU hours required for model training, this optimization renders our method negligible in computational cost, especially considering it is a one-time preprocessing step preceding the training of generative models.

As an effective alternative to the k-means clustering algorithm in the codebook token feature space, our proposed DCPE is inherently plug-and-play and can be seamlessly integrated into existing codebook prior-based methods. To demonstrate the effectiveness of our DCPE, we integrated it into the pipelines of both CTF~\cite{guo2025improving} and IAR~\cite{hu2025improving} in our experiments. First, CTF trains an autoregressive model with coarse cluster indices to generate cluster index sequences, and a refinement model to convert these indices into real tokens. We observed that the limited number of clusters (set to 1/32 of the total token count) and the use of a suboptimal k-means algorithm led to clusters with diverse semantic tokens. This necessitated a large refinement model to achieve image generation. We suggest that by using DCPE with a relatively larger number of clusters, the refinement model could even be eliminated while still maintaining competitive results. The corresponding experimental results are shown in \cref{tab:main_reduced_codebook}.

IAR introduces an auxiliary cluster-oriented loss upon the basic autoregressive next-token cross-entropy loss, accelerating training by predicting the correct cluster index. Integrating DCPE into this framework is straightforward. Besides, IAR uses balanced k-means clustering to ensure that the cluster sizes are equal, allowing the direct comparison of different cluster logits, which are defined as the sum of all token logits in a cluster. However, as mentioned earlier regarding the token space disparity issue, this approach does not align with the actual semantic distribution of tokens, and DCPE does not impose such a restriction on cluster size in its design. We investigate this difference in \cref{fig:cluster_size}. To make the cluster logits compatible with DCPE in IAR, we redefine the cluster logits as the mean of the logits of all tokens within that cluster. The experimental results can be found in \cref{tab:main_IAR}.

\section{Experiments}

\subsection{Implementation Details}
\label{sec:impl_details}

\subsubsection{Model and Training} 
Following existing codebook prior-based methods~\cite{hu2025improving,guo2025improving}, we adopt LlamaGen~\cite{sun2024autoregressive} as our backbone and utilize its off-the-shelf image tokenizer. This tokenizer has a codebook size of 16,384 and downsamples the input image by a factor of 16$\times$16. For the training of the autoregressive model, we strictly follow the original settings of LlamaGen. In order to train models with a clustered codebook, we convert the token indices extracted from the training dataset into the corresponding cluster indices based on the clustering results. Since the generative model trained in this way can only predict cluster indices, we adopt a naive random selection strategy to decode the image, randomly selecting a token from the predicted cluster. Unless otherwise specified, experiments marked with ``+ DCPE'' follow this decoding strategy.

\begin{algorithm}[t]
\caption{PyTorch Pseudo-Code for Our DCPE}
\label{algo:pseduo_code}
\begin{algorithmic}[0]
\footnotesize
\STATE \mycomment{\texttt{\# codebook: size [N, d]}}
\STATE \mycomment{\texttt{\# k: target number of clusters}}
\STATE \texttt{dist = torch.cdist(codebook, codebook, p=2)}
\STATE \texttt{dist.fill\_diagonal\_(float(\textquotesingle inf\textquotesingle))}
\STATE \texttt{sizes = torch.ones(N)}
\STATE \texttt{labels = torch.arange(N)}
\STATE \mykeyword{\texttt{for}} \texttt{m} \mykeyword{\texttt{in}} \texttt{range(N, k, -1):}
\STATE \quad \mycomment{\texttt{\# get position of nearest clusters}}
\STATE \quad \texttt{min\_pos = torch.argmin(dist.view(-1), dim=0)}
\STATE \quad \texttt{min\_pos\_i, min\_pos\_j = min\_pos//m, min\_pos\%m}
\STATE \quad \mycomment{\texttt{\# get real cluster index since \textit{dist} is merged}}
\STATE \quad \texttt{label\_i = get\_real\_label(min\_pos\_i)}
\STATE \quad \texttt{label\_j = get\_real\_label(min\_pos\_j)}
\STATE \quad \mycomment{\texttt{\# update cluster index label}}
\STATE \quad \texttt{labels[labels == label\_j] = label\_i}
\STATE \quad \mycomment{\texttt{\# add j-th row/column to i-th row/column}}
\STATE \quad \texttt{sizes = add\_j\_to\_i(sizes,min\_pos\_i,min\_pos\_j)}
\STATE \quad \texttt{dist = add\_j\_to\_i(dist, min\_pos\_i, min\_pos\_j)}
\STATE \quad \mycomment{\texttt{\# remove j-th row and column}}
\STATE \quad \texttt{sizes = remove\_j(sizes, min\_pos\_j)}
\STATE \quad \texttt{dist = remove\_j(dist, min\_pos\_j)}
\end{algorithmic}
\end{algorithm}

For training the refine model, we follow the training configuration of the vanilla LlamaGen, with the only modification being a reduction of the training epochs to 100. In terms of model structure, we build upon the LlamaGen-B codebase and remove the causal mask, consistent with the approach described in CTF~\cite{guo2025improving}. Additionally, as shown in \cref{tab:refine_layer}, we reduce the number of model layers to one due to our improved codebook prior. Under this configuration, the training time of our refine model is approximately 10\% of that of the refine model in CTF, significantly improving training efficiency.

For experiments integrating DCPE into IAR, we set the number of clusters to 512 and the cluster loss weight to 1.0, keeping all other training settings consistent with the original IAR using our reproduced code. Moreover, due to the mean operation discussed in \cref{sec:impl_anal}, the gradients of tokens from different clusters are imbalanced. This gradient imbalance can lead to performance degradation when cluster sizes are highly skewed. To alleviate this, we simply prevent the merging of clusters whose sizes exceed $N/k$. While this serves as a temporary solution, a deeper investigation into this issue is left for future work. More detailed hyperparameter settings can be found in the appendix.

\subsubsection{Evaluation}
Following LlamaGen~\cite{sun2024autoregressive}, we evaluate the image generation performance on the class-conditional image generation task using the ImageNet-1K benchmark~\cite{deng2009imagenet}. The generated images have a resolution of 256$\times$256, corresponding to a token sequence length of 256. For each evaluation, we generate a total of 50,000 images across all 1,000 classes. We compute the FID, Inception Score (IS), precision (Pr.), and recall (Re.) using the same code as existing works~\cite{sun2024autoregressive}. rFID, PSNR, and SSIM are also included to evaluate reconstruction performance. 
FID measures the distribution similarity between the features of training images and generated images; a lower FID indicates better generation quality. IS evaluates image quality and diversity by computing the entropy of features extracted from generated images; a higher IS reflects better quality and diversity. Precision and Recall assess class-conditional generation performance, where higher values denote greater accuracy. rFID compares the reconstructed image with the raw image, with lower values also suggesting better reconstruction. PSNR quantifies pixel-level differences between two images, while SSIM evaluates perceptual similarity. Higher PSNR and SSIM scores indicate better perceptual alignment between the reconstructed and raw images.
Additional details on generation hyperparameters can be found in the appendix.

\subsection{Main Results}
\label{sec:main_res}

\subsubsection{Training with Reduced Codebook} 
As discussed in \cref{sec:impl_anal}, we train autoregressive image generation models using cluster indices obtained from different clustering algorithms. We compare the commonly used k-means method with our proposed DCPE approach, as shown in \cref{tab:main_reduced_codebook}. By applying clustering, we reduce the codebook size by half and adopt the inference strategy described in \cref{sec:impl_details}. This vocabulary reduction decreases the number of parameters in the input and output layers of the autoregressive models. From the numerical results, it is evident that models trained with k-means clustering experience a performance drop compared to the baseline. This degradation is likely due to k-means’ inability to capture token similarity information, hindering the model’s ability to fit the training data. In contrast, our DCPE approach yields better results by exploiting a discriminative codebook prior. Notably, in experiments on LlamaGen-B, our method outperforms the baseline despite using fewer parameters and a smaller vocabulary. We attribute this to the difficulty small autoregressive models face in fitting complex token distributions, while the codebook prior extracted by DCPE makes convergence easier. On larger models, our approach also maintains performance comparable to the baseline.

\begin{table*}[t]
\centering
\caption{Performance of vanilla and cluster-based LlamaGen~\cite{sun2024autoregressive} on class-conditional ImageNet at 256$\times$256 resolution.
}
\resizebox{0.9\textwidth}{!}{
\begin{tabular}{l P{15mm} P{17mm}|P{14mm} P{14mm} P{14mm} P{14mm}}
% \begin{tabular}{lcc|cccc
\toprule
Method & \#Para. & \#Vocab. & FID↓ & IS↑  & Precision↑ & Recall↑ \\ \midrule
LlamaGen-B$^\dagger$ \cite{sun2024autoregressive}  & 111M & 16384 & 5.29   &   185.7    & 0.84     &   0.45 \\
\quad + k-means & 94M & 8192   & 5.56   &   188.4    & 0.83     &   0.44    \\
\rowcolor{mygray} \quad + DCPE & 94M & 8192   & 4.83  & 198.8    & 0.82     & 0.47    \\
\midrule
LlamaGen-L$^\dagger$ \cite{sun2024autoregressive}  & 343M & 16384   & 3.68   &   248.9    & 0.83     &   0.52    \\
\quad + k-means & 311M & 8192   & 4.08   &   216.7    & 0.79     &   0.55    \\
\rowcolor{mygray} \quad + DCPE & 311M & 8192  & 3.34   & 238.4    & 0.81     & 0.54      \\
\midrule
LlamaGen-XL$^\dagger$ \cite{sun2024autoregressive}  & 775M & 16384   & 3.14   &   269.9    & 0.83     &   0.54    \\
\quad + k-means & 719M & 8192   & 3.32   &   240.1    & 0.79     &   0.58    \\
\rowcolor{mygray} \quad + DCPE & 719M & 8192   & 2.86   & 267.4    & 0.82     & 0.56      \\ \bottomrule
\end{tabular}
}

\vspace{1mm}

\makebox[\textwidth][c]{
\parbox{0.88\textwidth}{
\textbf{$^\dagger$} indicates results reproduced using the official code.
``\#Vocab.'' denotes the codebook size for the baseline and the number of clusters for cluster-based methods.
Our DCPE achieves comparable or even better performance than the baseline using fewer parameters and a smaller vocabulary size, especially on smaller models, while k-means shows degradation due to its inferior codebook prior.
}
}
\label{tab:main_reduced_codebook}
\end{table*}

\begin{figure}[t]
  \centering
  \includegraphics[width=0.85\columnwidth]{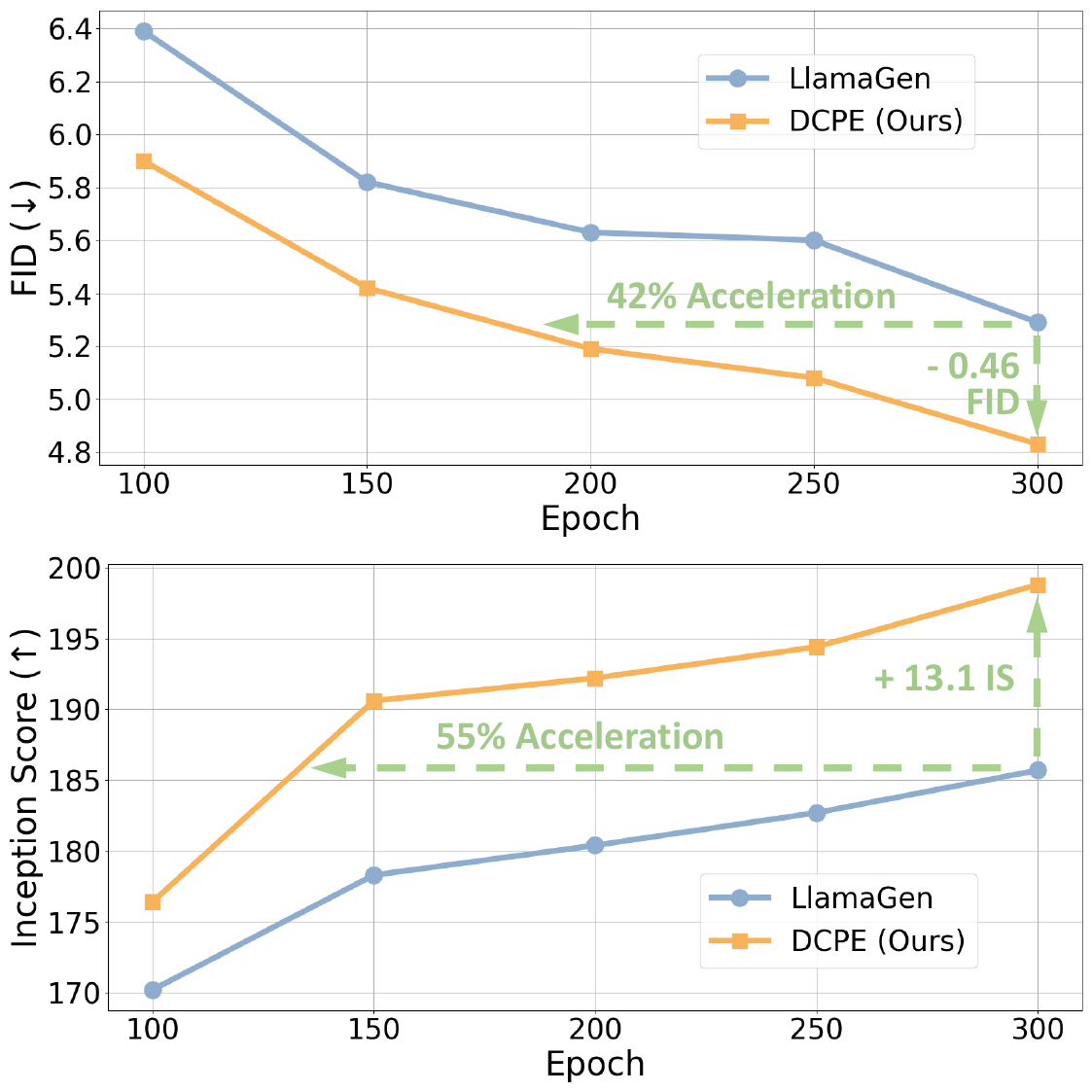} 
  \caption{Performances at different epochs of the vanilla LlamaGen-B and its cluster-based training with our DCPE. The codebook prior extracted by DCPE can effectively accelerate the models' convergence.}
    \label{fig:results_per_epoch}
\end{figure}

To demonstrate the convergence acceleration effect mentioned above, we compare the performance of the vanilla and cluster-based LlamaGen-B across different epochs, as shown in \cref{fig:results_per_epoch}. The results indicate that training with clusters obtained by DCPE significantly accelerates the convergence of the autoregressive model. For example, to reach the same FID and IS as the baseline, training with DCPE achieves speedups of 42\% and 55\%, respectively. When trained for the same number of epochs, our method reduces the FID by 0.46 and improves the IS by 13.1.

While CTF~\cite{guo2025improving} requires training a large refinement model (311M parameters) due to its suboptimal use of k-means, our DCPE performs well even without a refinement model as shown in \cref{tab:main_reduced_codebook}. We further conducted experiments by training a refinement model, similar to CTF, to map predicted clusters to real tokens. However, benefiting from the strong baseline performance of our method, we only train a small one-layer attention network unlike CTF. As shown in \cref{tab:main_refine}, with a small refinement model of 25M parameters, we achieve further performance improvements, particularly reflected in the increased IS score, which indicates better image visual quality. Nevertheless, we observe a slight degradation in FID, which we attribute to distribution shifts caused by the refinement model, since FID measures the divergence between the generated and training image distributions. Notably, on LlamaGen-L and LlamaGen-XL, the results with the refinement model begin to surpass the performance of vanilla training (\cref{tab:main_reduced_codebook}), despite using fewer parameters and a smaller vocabulary. We also trained refinement models with fewer clusters (see \cref{sec:abl}), and found that such a small one-layer network performs well under a large codebook downsampling rate.

To further demonstrate that DCPE can provide improvements across different image tokenizers with varying token distributions, we replaced the original image tokenizer with the larger GigaTok~\cite{xiong2025gigatok}, as shown in \cref{tab:main_gigatok}. Due to the tokenizer being scaled up significantly (10$\times$ parameters), a reduced codebook can no longer improve performance due to missing details. However, the results still show that DCPE outperforms k-means across different vocabulary sizes. Notably, the performance gap becomes more significant as the vocabulary size decreases. This suggests that with larger clusters, k-means is more likely to produce clusters containing tokens with different semantics. In contrast, the agglomerative design of DCPE better preserves semantic consistency in each cluster.

\subsubsection{Plug-and-play Integration to IAR} We integrate our DCPE into the IAR pipeline as a replacement for the default k-means clustering. The implementation details are described in \cref{sec:impl_anal} and \ref{sec:impl_details}. As shown in \cref{tab:main_IAR}, replacing k-means clustering with our DCPE effectively improves the performance of IAR. This improvement can be attributed to DCPE’s ability to extract a more discriminative codebook prior, which provides better auxiliary information for the training of autoregressive generation model.

\begin{table}[t]
\centering
\caption{Performance of a small refine model on DCPE.
}
\resizebox{\columnwidth}{!}{
    \begin{tabular}{l P{14mm}|P{7mm} P{7mm} P{6mm} P{6mm}}
    \toprule
    Method & \#Para. & FID↓ & IS↑  & Pr.↑ & Re.↑ \\ \midrule
    DCPE & 94M    & 4.83  & 198.8    & 0.82     & 0.47    \\
    \rowcolor{mygray} \quad + refine & 94+25M    & 5.02  & 204.5    & 0.84     & 0.45    \\
    \midrule
    DCPE & 311M   & 3.34   & 238.4    & 0.81     & 0.54      \\
    \rowcolor{mygray} \quad + refine & 311+25M   & 3.41   & 248.7    & 0.83     & 0.52      \\
    \midrule
    DCPE & 719M    & 2.86   & 267.4    & 0.82     & 0.56      \\ 
    \rowcolor{mygray} \quad + refine & 719+25M    & 3.11   & 273.4    & 0.83     & 0.54      \\ \bottomrule
    \end{tabular}
}

\vspace{1mm}

\makebox[\columnwidth][c]{
\parbox{0.98\columnwidth}{
Evaluations are conducted under the same settings as in \cref{tab:main_reduced_codebook}. ``+ refine'' means a small one-layer attention network is trained to refine the predicted clusters into tokens, similar to what CTF~\cite{guo2025improving} does. Explanations of the performance gap can be found in \cref{sec:main_res}.
}
}
\label{tab:main_refine}
\end{table}

\begin{table}[t]
\centering
\caption{Performance of DCPE with image tokenizer GigaTok-B-L~\cite{xiong2025gigatok}. 
}
\resizebox{\columnwidth}{!}{
    \begin{tabular}{lc|cccc}
    \toprule
    Method & \#Vocab. & FID↓ & IS↑  & Pr.↑ & Re.↑ \\ \midrule
    GigaTok$^\dagger$ \cite{xiong2025gigatok} & 16384 & 3.39   &   263.7    & 0.81     &   0.55 \\ \midrule
    \quad + k-means & 8192  & 3.67   &   259.8    & 0.79     &   0.56    \\
    \rowcolor{mygray} \quad + DCPE & 8192 & 3.65  & 261.5    & 0.80     & 0.55    \\ \midrule
    
    \quad + k-means & 4096  & 4.74   &   240.2    & 0.77     &   0.55    \\
    \rowcolor{mygray} \quad + DCPE  & 4096  & 4.19   & 249.3    & 0.78     & 0.56      \\ \bottomrule
    \end{tabular}
}

\vspace{1mm}

\makebox[\columnwidth][c]{
\parbox{0.98\columnwidth}{
Evaluations are conducted under the same settings as in \cref{tab:main_reduced_codebook}.
\textbf{$^\dagger$} indicates results reproduced using the official code.
Our DCPE consistently outperforms k-means clustering across different vocabulary sizes.
}
}
\label{tab:main_gigatok}
\end{table}

\begin{table}[t]
\centering
\caption{Performance comparison of integrating DCPE to IAR on class-conditional ImageNet at 256$\times$256 resolution.
}
% \vspace{3mm}
\resizebox{\columnwidth}{!}{
\begin{tabular}{c|lc|cccc}
\toprule
Type                    & Model                        & \#Para.                   & FID↓                         & IS↑                            & Pr.↑                         & Re.↑                         \\ \midrule
                        & BigGAN \cite{brock2018large}                      & 112M                      & 6.95                         & 224.5                          & 0.89                         & 0.38                         \\
                        & GigaGAN \cite{kang2023scaling}                     & 569M                      & 3.45                         & 225.5                          & 0.84                         & 0.61                         \\
\multirow{-3}{*}{GAN}   & StyleGAN-XL \cite{sauer2022stylegan}                 & 166M                      & 2.30                          & 265.1                          & 0.78                         & 0.53                         \\ \midrule
                        & ADM \cite{dhariwal2021diffusion}                         & 554M                      & 10.94                        & 101.0                            & 0.69                         & 0.63                         \\
                        & CDM \cite{ho2022cascaded}                         & -                         & 4.88                         & 158.7                          & -                            & -                            \\
                        & LDM-4 \cite{rombach2022high}                       & 400M                      & 3.60                          & 247.7                          & -                            & -                            \\
\multirow{-4}{*}{Diff.} & DiT-XL/2 \cite{peebles2023scalable}                    & 675M                      & 2.27                         & 278.2                          & 0.83                         & 0.57                         \\ \midrule
                        & MaskGIT \cite{chang2022maskgit}                     & 227M                      & 6.18                         & 182.1                          & 0.80                          & 0.51                         \\
\multirow{-2}{*}{Mask.} & MaskGIT-re \cite{chang2022maskgit}                  & 227M                      & 4.02                         & 355.6                          & -                            & -                            \\ \midrule
                        & VAR-d16 \cite{tian2024visual}                     & 310M                      & 3.30                          & 274.4                          & 0.84                         & 0.51                         \\
                        & VAR-d20 \cite{tian2024visual}                     & 600M                      & 2.57                         & 302.6                          & 0.83                         & 0.56                         \\
\multirow{-3}{*}{VAR}   & VAR-d24 \cite{tian2024visual}           & 1.0B                      & 2.09                         & 312.9                          & 0.82                         & 0.59                         \\ \midrule
                        & VQGAN \cite{esser2021taming}                      & 227M                      & 18.65                        & 80.4                           & 0.78                         & 0.26                         \\
                        & VQGAN \cite{esser2021taming}                       & 1.4B                      & 15.78                        & 74.3                           & -                            & -                            \\
                        & VQGAN-re \cite{esser2021taming}                    & 1.4B                      & 5.20                          & 280.3                          & -                            & -                            \\
                        & ViT-VQGAN \cite{yu2021vector}                   & 1.7B                      & 4.17                         & 175.1                          & -                            & -                            \\
                        & ViT-VQGAN-re \cite{yu2021vector}                & 1.7B                      & 3.48                         & 175.1                          & -                            & -                            \\
                        & RQTran. \cite{lee2022autoregressive}                     & 3.8B                      & 7.55                         & 134.0                            & -                            & -                            \\

                        & RQTran.-re \cite{lee2022autoregressive}                  & 3.8B                      & 3.80                          & 323.7                          & -                            & -                            \\
                        & LlamaGen-B$^\dagger$ \cite{sun2024autoregressive}  & 111M   & 5.29   &   185.7    & 0.84     &   0.45    \\
                        & LlamaGen-L$^\dagger$ \cite{sun2024autoregressive}  & 343M   & 3.68   &   248.9    & 0.83     &   0.52    \\
\multirow{-10}{*}{AR}   & LlamaGen-XL$^\dagger$ \cite{sun2024autoregressive}  & 775M   & 3.14   &   269.9    & 0.83     &   0.54    \\ \midrule
                        & IAR-B \cite{hu2025improving} & 111M   & 5.14 & 202.0 & 0.85 & 0.45   \\
                        & \cellcolor{mygray}\quad + DCPE & \cellcolor{mygray}111M   & \cellcolor{mygray}5.12 &   \cellcolor{mygray}209.5    & \cellcolor{mygray}0.86     & \cellcolor{mygray}0.44    \\
                        & IAR-L \cite{hu2025improving} & 343M   & 3.18 & 234.8 & 0.82 & 0.53    \\
                        & \cellcolor{mygray}\quad + DCPE & \cellcolor{mygray}343M   & \cellcolor{mygray}3.14   & \cellcolor{mygray}249.5    & \cellcolor{mygray}0.83     & \cellcolor{mygray}0.53      \\
                        & IAR-XL \cite{hu2025improving} & 775M   & 2.52 & 248.1 & 0.82 & 0.58    \\
\multirow{-6}{*}{IAR} & \cellcolor{mygray}\quad + DCPE & \cellcolor{mygray}775M   & \cellcolor{mygray}2.49   & \cellcolor{mygray}270.8    & \cellcolor{mygray}0.83     & \cellcolor{mygray}0.58 \\ \bottomrule
\end{tabular}
}

\vspace{1mm}

\makebox[\columnwidth][c]{
\parbox{0.98\columnwidth}{
\textbf{$^\dagger$} indicates results reproduced using the official code.
Replacing the default k-means clustering with our DCPE improves the performance of IAR.
}
}
\label{tab:main_IAR}
\end{table}

\subsection{Ablation Study and Analysis}
\label{sec:abl}

\subsubsection{Ablation Study on Algorithm Designs} 
As discussed in \cref{sec:prelim}, we attribute the suboptimality of k-means in extracting information from the token feature space to two key issues: token space disparity and centroid distance inaccuracy. To address these issues, we conduct experiments by adopting agglomerative clustering and a centroid-free distance calculation, respectively, as shown in \cref{tab:diff_cluster_algo}. In addition to the baseline k-means and its common variant k-means++, we independently evaluate the effects of agglomerative clustering and the centroid-free distance calculation. To isolate the impact of agglomerative clustering, we remove the use of instance-based distance in our DCPE. Conversely, to test the centroid-free distance calculation in isolation, we incorporate it into the naive k-means algorithm with other components unmodified. The results show that both techniques contribute to more effective extraction of the codebook prior, leading to improved performance of the autoregressive model. While density-based clustering methods such as DBSCAN~\cite{ester1996density} may also alleviate the issues mentioned above, their inability to control the number of clusters makes them impractical for our tasks; thus we do not include them in our experiments.

\subsubsection{Token Similarity in Clusters} The effectiveness of our DCPE in enhancing the token semantic similarity within a cluster, as discussed in \cref{sec:DCPE}, is further validated through a reconstruction experiment. Here we replace all image tokens obtained by the LlamaGen tokenizer~\cite{sun2024autoregressive} with tokens from the same cluster, and then pass them into the decoder to reconstruct the image. 
Specifically, we select 50,000 images from the ImageNet validation set and apply center cropping to them. The reconstructed images are then compared with the original ones by computing rFID, PSNR, and SSIM~\cite{sun2024autoregressive,peebles2023scalable} under different clustering methods. Since the official LlamaGen codebase does not provide implementations for PSNR and SSIM evaluation, we follow common practice and utilize the corresponding functions from the Python package scikit-image for evaluation. Using the open-source LlamaGen image tokenizer along with our evaluation implementation, we are able to reproduce the rFID and PSNR values reported in the LlamaGen paper. However, the SSIM results show discrepancies compared to the original report. To ensure a fair comparison, we report our reproduced results.

The results in \cref{tab:recon} show that using clusters generated by our DCPE leads to less degradation in image quality compared to k-means. This advantage becomes even more pronounced as the vocabulary size decreases, i.e., the size of clusters increases. The results indicate that our DCPE better preserves semantic consistency within clusters, allowing perturbed image tokens to be decoded correctly.

\begin{table}[t]
\centering
\caption{Ablation study on two proposed designs.}
\resizebox{\columnwidth}{!}{
\begin{tabular}{l|cc|cc}
    \toprule
    Method & C-F & Agglo. & FID↓ & IS↑ \\ \midrule
    k-means & $\times$ & $\times$ & 5.56 & 188.4 \\
    k-means++ & $\times$ & $\times$ & 5.33 & 187.9 \\
    {\small DCPE w/ centroi}d & $\times$ & $\checkmark$ & 5.25 & 197.5 \\
    {\small k-means w/o centroid} & $\checkmark$ & $\times$ & 5.01 & 189.3 \\
    \rowcolor{mygray} DCPE & $\checkmark$ & $\checkmark$ & 4.83 & 198.8 \\
    \bottomrule
\end{tabular}
}

\vspace{1mm}

\makebox[\columnwidth][c]{
\parbox{0.98\columnwidth}{
Evaluations are conducted under the same settings as in \cref{tab:main_reduced_codebook}.
``C-F'' means centroid-free and ``Agglo.'' means agglomerative clustering.
% The two main components of our DCPE, agglomerative clustering and centroid-free distance, both enhance the extraction of the codebook prior, boosting the generation performance.
Both two main components of our DCPE contribute to improved generation performance.
}
}
\label{tab:diff_cluster_algo}
\end{table}

\subsubsection{Cluster Size Analysis} 
\begin{figure}[t]
    \centering
    \includegraphics[width=0.9\linewidth]{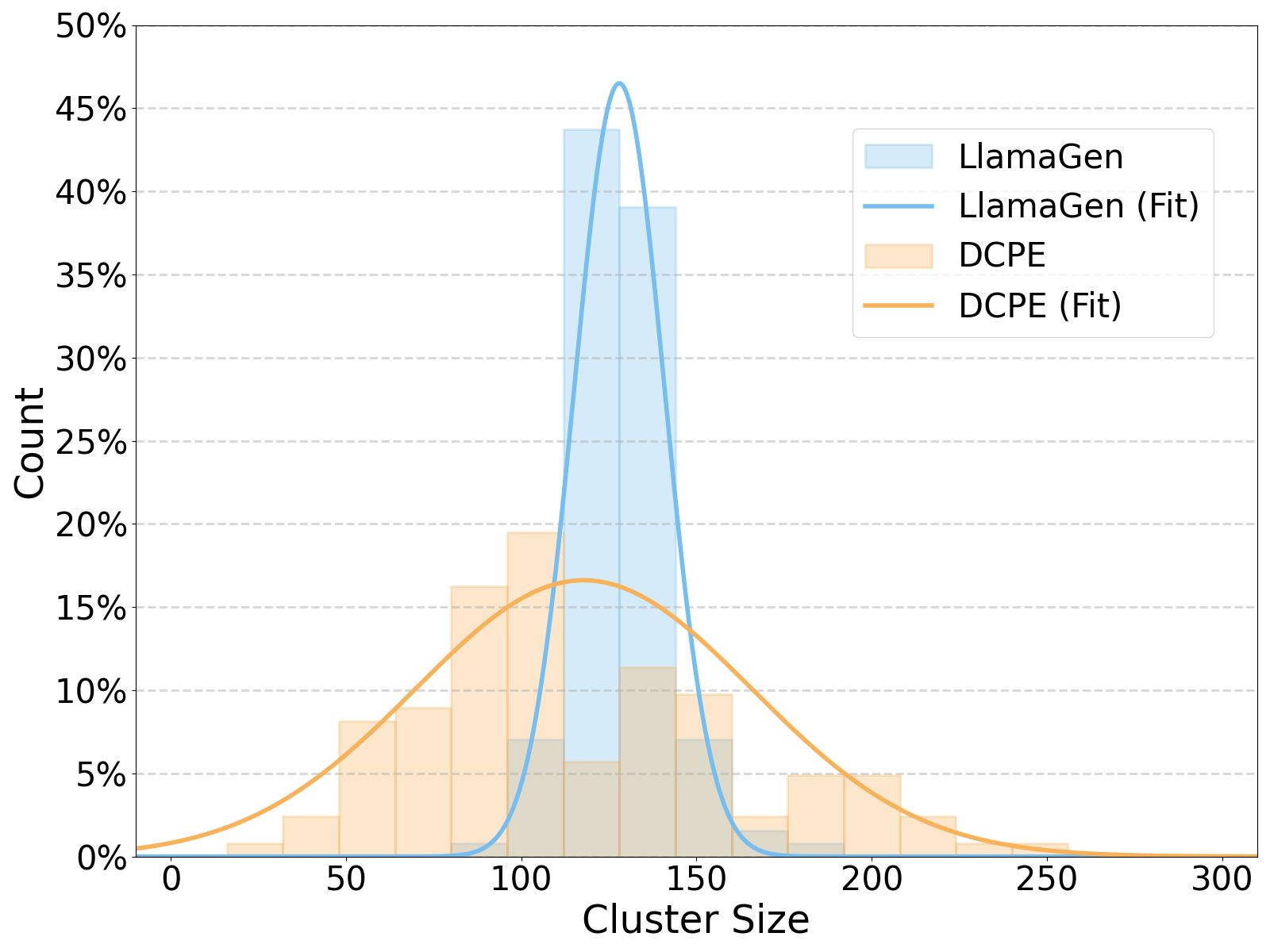} 
    \caption{Cluster sizes obtained by k-means clustering and our DCPE. We derive 128 clusters from the 16k-sized codebook and use a Gaussian distribution to fit them for better visualization. Our DCPE produces clusters with a wider range of sizes, which is consistent with the inherent sparsity of token density.}
    \label{fig:cluster_size}
\end{figure}

To demonstrate that our DCPE effectively handles non-uniformly distributed token feature spaces, we analyzed the distribution of cluster sizes generated by both k-means clustering and our DCPE, as shown in \cref{fig:cluster_size}. For better visualization, we clustered a 16k-sized codebook used in LlamaGen into 128 clusters. The results indicate that clusters formed by k-means exhibit relatively uniform sizes, which contradict the intrinsic sparsity of the token feature space discussed in \cref{sec:prelim}. In contrast, DCPE produces clusters with a wider range of sizes. This variation is attributed to our agglomerative clustering strategy, which prioritizes merging similar tokens to ensure intra-cluster similarity and avoids the synchronous updates in k-means clustering, which tend to overlook variations in density.

\subsubsection{Performance under Different Vocabulary Sizes} We evaluate our DCPE under different vocabulary sizes, i.e. the number of clusters, and train a refine model for each setting. The results are shown in \cref{tab:diff_rate}. Since smaller vocabulary sizes require fewer input and output layer parameters, the number of model parameters, reported in the ``\#Para.'' column, varies accordingly. As shown by the results, models trained with smaller vocabulary sizes tend to perform worse. This degradation occurs because tokens grouped into the same cluster become semantically more diverse as the cluster size increases, leading to a greater loss of detail in the generated images. However, we found that introducing a lightweight refine model, composed of a one-layer attention mechanism, can significantly improve performance. For example, with a vocabulary size of 2048, the refine model achieves comparable performance to the default setting with fewer parameters. Nonetheless, when the vocabulary size becomes too small, such as 1024, the excessive semantic mixing within clusters increases the convergence difficulty of generative models.

\begin{table}[t]
\centering
\caption{Image reconstruction performance with replaced tokens.}
\resizebox{\columnwidth}{!}{
\begin{tabular}{l P{9mm}|P{9mm} P{9mm} P{9mm}}
    \toprule
    Method & \#Vocab. & rFID↓ & PSNR↑ & SSIM↑ \\ \midrule
    w/o replacement$^\dagger$ & 16384 & 2.20   &  20.69 & 0.514  \\ \midrule
    \quad w/ k-means & 8192  & 4.14   &   19.57    & 0.471    \\
    \rowcolor{mygray} \quad w/ DCPE & 8192 & 3.48  & 19.78    & 0.479  \\ \midrule
    
    \quad w/ k-means & 4096  & 8.27  & 18.90    & 0.442     \\
    \rowcolor{mygray} \quad w/ DCPE  & 4096  & 7.23  & 19.01    & 0.447  \\ \bottomrule
\end{tabular}
}

\vspace{1mm}

\makebox[\columnwidth][c]{
\parbox{0.98\columnwidth}{
We replace image tokens extracted by the LlamaGen tokenizer with random tokens from the same cluster (see \cref{sec:abl}).
\textbf{$^\dagger$} indicates results reproduced with the official code.
Using clusters from our DCPE shows less performance degradation due to consistent token semantics in each cluster.
}
}
\label{tab:recon}
\end{table}

\subsubsection{Performance with Larger Refine Model} In \cref{sec:main_res}, we discussed that, unlike CTF, which employs a large refine model, we train only a one-layer attention network, benefiting from the strong baseline performance provided by our DCPE. We further conducted experiments to investigate whether using a larger refine model could improve the performance of the cluster-based generation model. As shown in \cref{tab:refine_layer}, increasing the size of the refine model yields only marginal performance gains. This is because, under our setting (vocabulary size = 8192), tokens within each cluster exhibit high similarity, and random selection is sufficient. Thus, refining cluster indices to token indices brings limited benefits.

\begin{table*}[t]
\centering
\caption{Performance of our DCPE under different vocabulary sizes on LlamaGen-B.}
\resizebox{0.8\textwidth}{!}{
\begin{tabular}{l P{18mm} P{18mm}|P{14mm} P{14mm} P{14mm} P{14mm}}
\toprule
Method & \#Para. & \#Vocab. & FID↓ & IS↑  & Precision↑ & Recall↑ \\ \midrule
LlamaGen-B$^\dagger$ \cite{sun2024autoregressive}  & 111M & 16384 & 5.29   &   185.7    & 0.84     &   0.45 \\ \midrule
\rowcolor {mygray}DCPE & 94M  & 8192   & 4.83  & 198.8    & 0.82     & 0.47    \\
\rowcolor {mygray}\quad + refine & 94M+25M & 8192    & 5.02  & 204.5    & 0.84     & 0.45    \\ \midrule
DCPE & 92M & 4096 & 5.52 & 191.8 & 0.80 & 0.48 \\
\quad + refine & 92M+24M & 4096 & 5.66 & 207.0 & 0.84 & 0.43 \\ \midrule
DCPE & 89M & 2048 & 9.44 & 161.5 & 0.73 & 0.50 \\ 
\quad + refine & 89M+22M & 2048 & 5.85 & 208.8 & 0.82 & 0.45 \\ \midrule
DCPE & 87M & 1024 & 18.70 & 111.0 & 0.63 & 0.49\\ 
\quad + refine & 87M+21M & 1024 & 7.21 & 193.5 & 0.80 & 0.41 \\ \bottomrule
\end{tabular}
}

\vspace{1mm}

\makebox[\textwidth][c]{
\parbox{0.78\textwidth}{
``+ refine'' means a small one-layer attention network is trained to refine the predicted clusters into tokens. Smaller vocabulary sizes tend to perform worse, while a small refine model can effectively improve performance, making a model with less parameter and smaller vocabulary perform comparably to the baseline.
}
}
% \vspace{-2mm}
\label{tab:diff_rate}
\end{table*}

\begin{table*}[t]
\centering
\caption{Performance of DCPE with different refine model sizes on LlamaGen-B.}
\resizebox{0.9\textwidth}{!}{
\begin{tabular}{l P{12mm} P{17mm} P{17mm}|P{14mm} P{14mm} P{14mm} P{14mm}}
\toprule
Method & \#Layers & \#Para. & \#Vocab. & FID↓ & IS↑  & Precision↑ & Recall↑ \\ \midrule
LlamaGen-B$^\dagger$ \cite{sun2024autoregressive} & \textbf{-} & 111M & 16384 & 5.29   &   185.7    & 0.84     &   0.45 \\ \midrule
\rowcolor{mygray}DCPE & \textbf{-} & 94M  & 8192   & 4.83  & 198.8    & 0.82     & 0.47    \\
\rowcolor{mygray}\quad + refine & 1 & 94M+25M & 8192    & 5.02  & 204.5    & 0.84     & 0.45    \\
\quad + refine & 3 & 94M+41M & 8192    & 5.18  & 204.7    & 0.84     & 0.45    \\
\quad + refine & 6 & 94M+62M & 8192    & 5.21  & 204.9    & 0.85     & 0.45    \\
\quad + refine & 9 & 94M+83M & 8192    & 5.20  & 205.8    & 0.85     & 0.45    \\
\quad + refine & 12 & 94M+105M & 8192    & 5.19  & 206.3   & 0.85     & 0.45    \\ \bottomrule
\end{tabular}
}

\vspace{1mm}

\makebox[\textwidth][c]{
\parbox{0.88\textwidth}{
``\#Layers'' indicates the number of attention layers in the refine model. Default settings are highlighted in gray. Scaling up the refine model leads to only marginal performance gains, because our DCPE generate high intra-similarity clusters, reducing the need for refinement.
}
}
% \vspace{-2mm}
\label{tab:refine_layer}
\end{table*}

\subsection{Hyperparameter Selection}

\subsubsection{Rationality of Random Selection in Inference} To validate the effectiveness of the random selection strategy used for properly inference with autoregressive models trained on cluster indices, we conducted experiments using different random selection seeds. We then evaluated the quality of the generated images to assess the stability of this generation paradigm. As shown in \cref{tab:refine_layer}, the evaluation metrics of the generated images remain consistent across different random seeds, with only small standard deviations observed. These results indicate that the tokens within each cluster produced by our DCPE exhibit high semantic similarity. Consequently, the randomly selected tokens convey similar semantic content, leading to minimal performance variation in the final decoded images.

\begin{figure}[t]
  \centering
  \includegraphics[width=0.85\columnwidth]{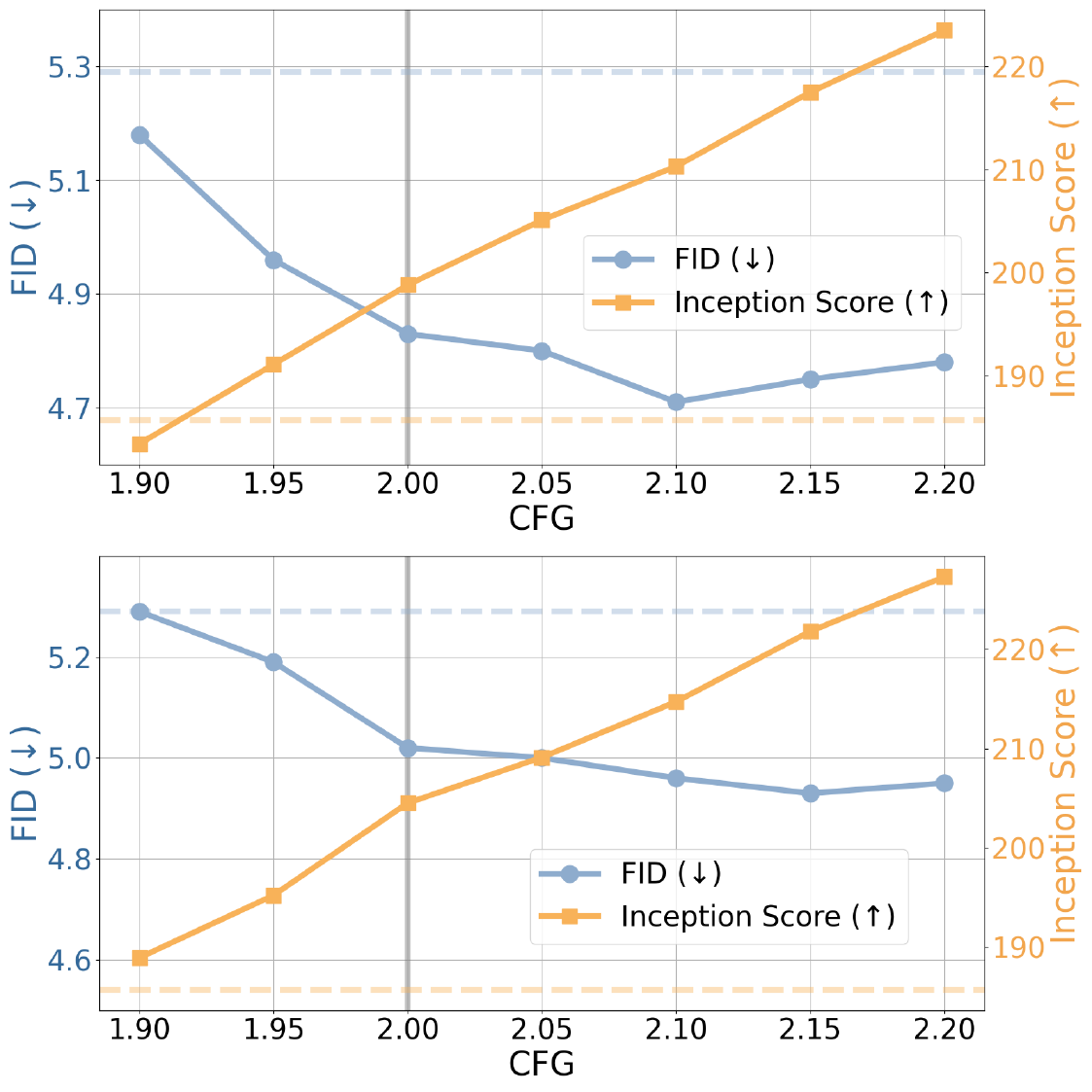} 
  \caption{Performance of DCPE (upper) and DCPE+Refine (lower) under different Classifier-Free Guidance (CFG) settings on LlamaGen-B. The dashed lines represent the results of vanilla LlamaGen-B. We select CFG = 2.0 as the default (highlighted with a gray line) due to its relatively balanced performance.}
  \label{fig:diff_cfg}
\end{figure}

\begin{figure*}[t]
    \centering
    \includegraphics[width=\textwidth]{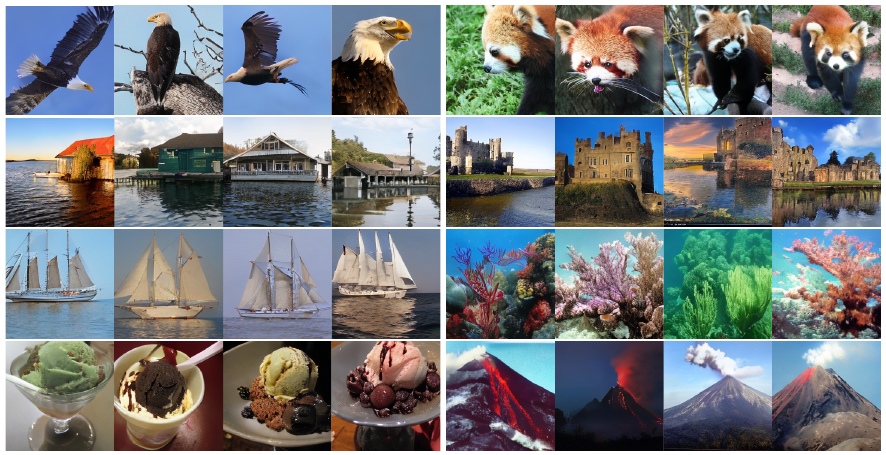} 
    % \vspace{-2mm}
    \caption{Generated images using LlamaGen-XL with our DCPE and a CFG of 4.0. More generated samples are provided in the appendix.}
    \label{fig:visual}
    % \vspace{-3mm}
\end{figure*}

\subsubsection{Classifier-Free Guidance} Similar to previous studies~\cite{sun2024autoregressive,tian2024visual,hu2025improving,guo2025improving}, we report generation metrics under different Classifier-Free Guidance (CFG) settings, as shown in \cref{fig:diff_cfg}. Consistent with existing findings, lower CFG values yield better FID scores, indicating closer alignment with the training data distribution. In contrast, higher CFG values result in higher inception scores, reflecting improved image quality. For a fair comparison, we follow prior works~\cite{sun2024autoregressive,hu2025improving,guo2025improving} and set the default CFG value to 2.0 for LlamaGen-B, which offers a favorable trade-off between FID and IS.

\subsubsection{Hyperparameters in IAR} As discussed in \cref{sec:impl_anal}, we integrat DCPE into the training framework of IAR~\cite{hu2025improving} and report superior results. We further perform an ablation study on two key hyperparameters in IAR: the number of clusters and the loss weight of the cluster-oriented cross-entropy loss. The results are shown in \cref{tab:IAR_hyper_para}, indicating that our default setting, 512 clusters and a loss weight of 1.0, yields relatively strong performance across different configurations.

\begin{table}[t]
\centering
\caption{Generation using different random selection seeds.}
\resizebox{\columnwidth}{!}{
\begin{tabular}{>{\arraybackslash}p{20mm} P{10mm} P{8mm}|P{10mm} P{10mm}}
\toprule
Method & \#Vocab. & Seed & FID↓ & IS↑  \\ \midrule
LlamaGen-B$^\dagger$ \cite{sun2024autoregressive}  & 16384 & 0  & 5.29   &   185.7 \\ \midrule
\multirow{5}{*}{DCPE}   & \multirow{5}{*}{8192}  & \cellcolor{mygray}0  & \cellcolor{mygray}4.83  & \cellcolor{mygray}198.8   \\
 &  & 1  & 4.92  & 199.1    \\
 &  & 2  & 4.93  & 195.4    \\
 &  & 3  & 4.84  & 197.6    \\
 &  & 4  & 4.84  & 198.7    \\ \midrule
Std. & \textbf{-} & \textbf{-} & 0.044 & 1.359\\ \bottomrule
\end{tabular}
}

\vspace{1mm}

\makebox[\columnwidth][c]{
\parbox{0.98\columnwidth}{
Evaluations are conducted under the same settings as in \cref{tab:main_reduced_codebook}.
The default settings are highlighted in gray. The results show minimal variation due to the high intra-similarity clusters generated by our DCPE.
}
}
\label{tab:diff_seed}
\end{table}

\begin{table}[t]
\centering
\caption{Ablation studies on hyperparameters used in IAR with DCPE.}
\resizebox{\columnwidth}{!}{
\begin{tabular}{P{10mm} P{10mm}|P{10mm} P{10mm} P{10mm} P{10mm}}
\toprule
\#Cluster & $\lambda$ & FID↓ & IS↑  & Precision↑ & Recall↑ \\ \midrule
\rowcolor{mygray} 512 & 1.0 & 5.12 & 209.5 & 0.86 & 0.44 \\ \midrule

\multirow{3}{*}{512}  & 0.2  & 5.15 & 202.6 & 0.85 & 0.45 \\
& 0.5  & 5.37 & 200.7 & 0.85 & 0.43 \\
& 2.0  & 5.33 & 198.2 & 0.86 & 0.43 \\ \midrule
128 & \multirow{3}{*}{1.0}  & 5.29 & 204.9 & 0.86 & 0.43  \\
256 & & 5.15 & 203.6 & 0.85 & 0.44 \\
1024 & & 5.23 & 202.3 & 0.85 & 0.45 \\ \bottomrule
\end{tabular}
}

\vspace{1mm}

\makebox[\columnwidth][c]{
\parbox{0.98\columnwidth}{
Evaluations are conducted under the same settings as LlamaGen-B in \cref{tab:main_IAR}.
``\#Cluster'' refers to the number of clusters, and $\lambda$ indicates the weight the cluster-oriented cross-entropy loss. Results obtained under the default setting are highlighted in gray.
}
}
\label{tab:IAR_hyper_para}
\end{table}

\subsection{Visualization Results}

We show the generated images using LlamaGen-XL with our DCPE and a CFG of 4.0 in \cref{fig:visual}, following the common setting~\cite{sun2024autoregressive,hu2025improving,guo2025improving}. The image generation classes include: bald eagle, lesser panda, boathouse, castle, schooner, coral reef, ice cream, and volcano. Additional generated samples are provided in the appendix.

\section{Limitation and Future Work}

Due to resource constraints, we conducted class-conditioned image generation experiments using only LlamaGen as the baseline. Recently, a growing number of autoregressive image generation paradigms have emerged, such as VAR~\cite{tian2024visual} with next-scale prediction, RAR~\cite{yu2024randomized} with randomized next-token prediction, and NAR~\cite{he2025neighboring} with next-neighbor prediction. Evaluating the effectiveness of our DCPE method on these emerging paradigms would also be valuable. Furthermore, text-conditioned image generation, which holds greater practical significance, deserves more extensive exploration.

As discussed in \cref{sec:impl_anal}, the clusters obtained by DCPE vary in size. To ensure correct training when integrating it into IAR~\cite{hu2025improving}, we redefine the cluster logits as the mean of the logits of all tokens within a cluster, instead of a naive summation. This modification introduces inconsistencies to the gradients of tokens within different clusters during backpropagation, which may lead to sub-optimal outcomes. We do not further investigate solutions to this issue, but a promising research direction would be to explore ways to enhance training stability~\cite{tang2024mind,tang2023source,he2025segment,he2024weakly,fang2025integrating}.

Regarding algorithm design, DCPE adopts a greedy strategy by merging the two nearest clusters at each step. This approach optimizes only the current iteration and does not guarantee a globally optimal solution. It may fail under specific token distributions. Given that finding a globally optimal clustering strategy is NP-hard, future work could explore more effective methods to approximate the global optimum.

\section{Conclusion}

In this work, we investigate and reveal the token space disparity and centroid distance inaccuracy problems in existing methods that leverage naive k-means clustering on codebook tokens to facilitate autoregressive image generation. 
To provide a better codebook prior, we present the Discriminative Codebook Prior Extractor (DCPE). It replaces unreliable centroid-based distances with more robust instance-based ones that reflect the true token similarities. Through an agglomerative merging strategy, DCPE further addresses token space disparity by preserving high-density regions.
Extensive experiments demonstrate that DCPE is plug-and-play and integrates seamlessly with existing codebook prior-based paradigms. By effectively utilizing the embedded token similarity, DCPE improves generation quality and accelerates training.

% \newpage
\bibliographystyle{IEEEtran}
\bibliography{my_bib}

\vfill

\end{document}